\documentclass[12pt]{article}
\usepackage{amsmath}
\usepackage{graphicx,psfrag,epsf}
\usepackage{enumerate}
\usepackage{natbib}
\usepackage{url} 
\usepackage{amsthm}
\usepackage{amssymb}
\usepackage{amsfonts}
\usepackage{bm}
\usepackage{latexsym}
\usepackage{setspace}
\usepackage{array}
\usepackage{float}
\usepackage{multirow}
\usepackage{soul, xcolor}
\usepackage[shortlabels]{enumitem}
\usepackage{algorithm}
\usepackage[noend]{algpseudocode}
\usepackage{afterpage}
\usepackage{lineno}

\newcommand{\blind}{0}

\begin{document}

\def\spacingset#1{\renewcommand{\baselinestretch}%
{#1}\small\normalsize} \spacingset{1}


\if0\blind
{
  \title{\bf Regression Phalanxes}
  \author{Hongyang Zhang \\
    Department of Statistics, University of British Columbia\\
    and \\
    William J. Welch \\
    Department of Statistics, University of British Columbia\\
    and \\
    Ruben H. Zamar \\
    Department of Statistics, University of British Columbia}
  \maketitle
} \fi

\if1\blind
{
  \bigskip
  \bigskip
  \bigskip
  \begin{center}
    {\LARGE\bf Title}
\end{center}
  \medskip
} \fi

\bigskip
\begin{abstract}
\citet{tomal2015ensembling} introduced the notion of ``phalanxes'' in the context of rare-class detection in two-class classification problems. A phalanx is a subset of features that work well for classification tasks. In this paper, we propose a different class of phalanxes for application in regression settings. We define a ``Regression Phalanx" -- a subset of features that work well together for prediction. We propose a novel algorithm which automatically chooses Regression Phalanxes from high-dimensional data sets using hierarchical clustering and builds a prediction model for each phalanx for further ensembling. Through extensive simulation studies and several real-life applications in various areas (including drug discovery, chemical analysis of spectra data, microarray analysis and climate projections) we show that an ensemble of Regression Phalanxes improves prediction accuracy when combined with effective prediction methods like Lasso or Random Forests.
\end{abstract}

\noindent%
{\it Keywords:} Regression Phalanxes, model ensembling, hierarchical clustering, Lasso, Random Forests
\vfill

\newpage
\spacingset{1.45} 

\section{Introduction}
\citet{tomal2015ensembling} introduced a novel approach for building diverse classification models, for the ensembling of classification models in the context of rare-class detection in two-class classification problems. They proposed an algorithm to divide the often large number of features (or explanatory variables) into subsets adaptively and build a base classifier (e.g. Random Forests) on each subset. The various classification models are then ensembled to produce one model, which ranks the new samples by their probabilities of belonging to the rare class of interest. The essence of the algorithm is to automatically choose the subset groups such that variables in the same group work well together for classification tasks; such groups are called phalanxes.

In this paper, we propose a different class of phalanxes for application in general regression tasks. We define a ``Regression Phalanx" -- a subset of features that work well together for regression (or prediction). We then propose a novel algorithm, with hierarchical clustering of features at its core, that automatically builds Regression Phalanxes from high-dimensional data sets and builds a regression model for each phalanx for further ensembling. 

In principle, any given regression method can be used as the base regression model. The goal is to improve the prediction accuracy of the base method. In this paper, we mainly focus on two well-established regression models: Lasso \citep{tibshirani1996regression} and Random Forests \citep{breiman2001random}. These two methods are known to have superior performance in various regression and prediction applications. For each application in this paper, we first compare the performances of Lasso and Random Forests (RF). The better performing method between the two is then chosen as the base regression model for building Regression Phalanxes.

The idea of ensembling Regression Phalanxes is promising because each Regression Phalanx is relatively low-dimensional. Thus, each variable makes a more significant contribution in the fitted model. Compared to training a full model where variables compete with each other in contributing to the final fit, more useful variables are likely to contribute to the ensembled regression model.

Our proposed phalanx-forming procedure resembles a hierarchical clustering of features (instead of samples), where ``similarity" between a pair of features (or subsets of features) is defined by how well they work together in the same regression or prediction model. With properly defined similarity measures, features can be then hierarchically merged into different phalanxes. 

The rest of paper is organized as follows. Section~2 presents the details of our proposed algorithm for building Regression Phalanxes. Section 3 presents a simple illustrative example, which forms the basis for simulation studies in Section 4. In Section~5, we demonstrate the performance of Regression Phalanxes on four additional real data sets. Finally, we conclude with some remarks and discussion of future work.

\section{Phalanx-formation algorithm}
In this section, the details of the Regression Phalanx formation algorithm are presented. The procedure is an agglomerative approach to build Regression Phalanxes, which is, in essence, a hierarchical clustering of variables. There are four key steps:
\begin{enumerate}
\item Initial grouping. Form $d$ initial groups from the original $D$ variables ($d \le D$).
\item Screening of initial groups. Screen out the underperforming initial groups to obtain $s \le d$ groups.
\item Hierarchical merging into phalanxes. Hierarchically merge the $s$ screened groups into $e \le s$ candidate phalanxes.
\item Screening of candidate phalanxes. Screen out the underperforming candidate phalanxes to obtain $h \le e$ final phalanxes.
\end{enumerate}

\begin{figure}[htbp]
\begin{center}
  \includegraphics[scale = 0.55]{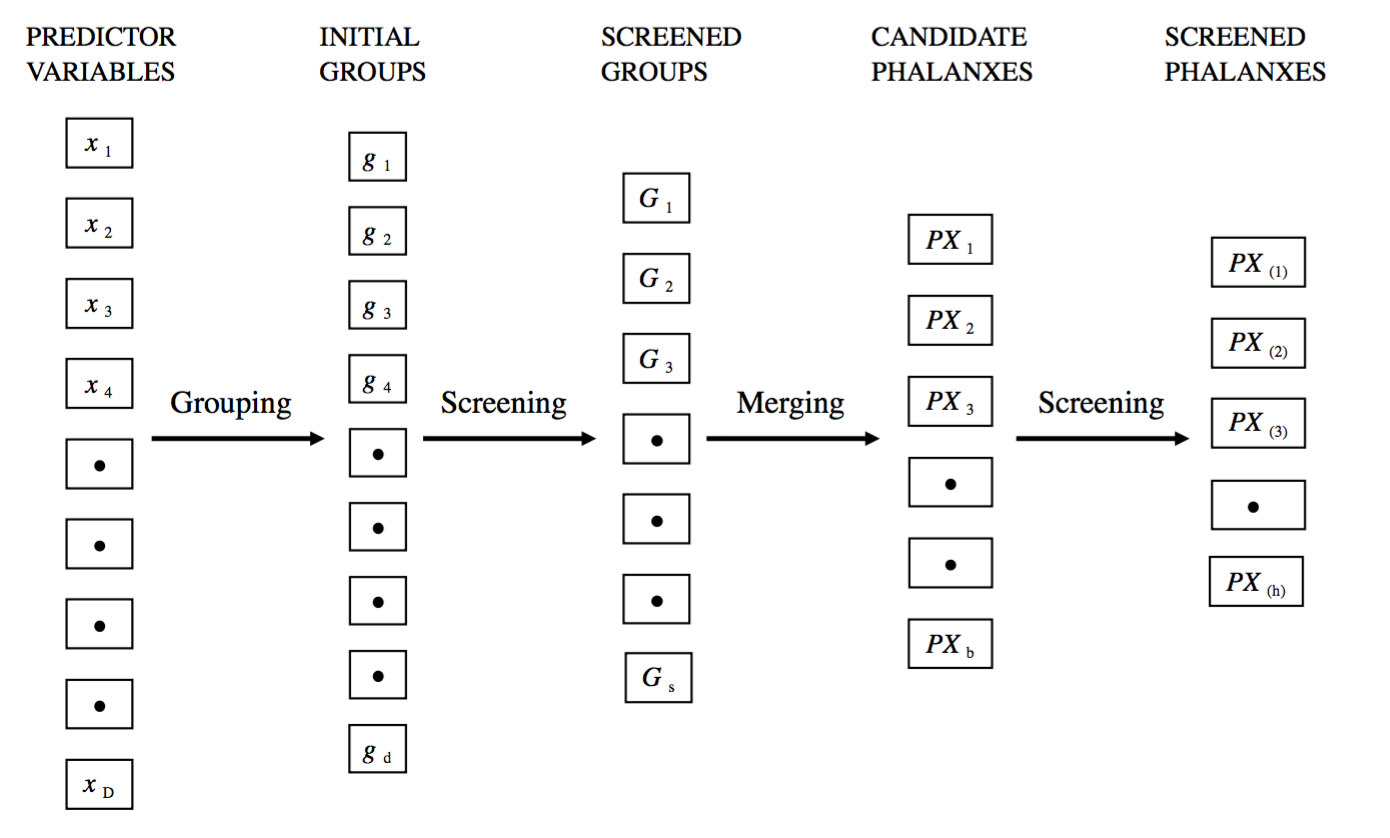}\\
  \end{center}
\caption{A sketch of the phalanx-formation procedure. $D$ variables are partitioned into $d$ initial groups, screened down to $s$ groups, combined into $e$ candidate phalanxes, and then screened down to $h$ phalanxes in the final ensemble ($D \ge d \ge s \ge e \ge h$).}
  \label{fig:sketch-phalanx}
\end{figure}

A sketch of the procedure is presented in Figure~\ref{fig:sketch-phalanx}. Each step of the phalanx-forming procedure is explained in more details in the following sections.

\subsection{Initial grouping}
This is an optional step. If this step is omitted, each initial group contains
a single  individual feature and the number of initial groups equals the
number of features. As a result, the following steps in the
phalanx-formation steps become more computational intensive since the
time complexity is quadratic in the number of groups. Thus, it is
recommended that the features be grouped into fewer initial groups if they lend themselves to natural
grouping (e.g.\ initial groups can be identified by features with
similar names). Also, 
if an initial group only contains a binary feature, its
corresponding model is likely to be weak since it can only
predict two possible values. On the other hand, an initial group with $k$
binary features can produce up to $2^k$ different  predictions. Thus, we 
recommend the grouping of the
 binary features into initial groups. If the data set contains a large number of features but no obvious hints for natural grouping, we can still use hierarchical clustering to obtain the initial groups. For binary features, \citet{tomal2015ensembling} proposed to use the Jaccard dissimilarity index, defined between binary features $\bm{x}_i$ and $\bm{x}_j$ as
\[d_J (\bm{x}_i, \bm{x}_j ) = 1-\frac{\bm{x}_i \cap \bm{x}_j}{\bm{x}_i \cup \bm{x}_j}.\]
Here $\bm{x}_i \cap \bm{x}_j$ is the number of observations where $\bm{x}_i$ and $\bm{x}_j$ both take the
value 1, and $\bm{x}_i \cup \bm{x}_j$ is the number of observations where $\bm{x}_i$ or $\bm{x}_j$ take the value 1. It is easy to see that  $0 \le d_J (\bm{x}_i, \bm{x}_j ) \le 1$. For continuous features (or a mix of binary and continuous features), we propose to use the ``1-Abs(Correlation)'' dissimilarity measure. That is, the dissimilarity between variable $x_i$ and $x_j$ is calculated as 
\[ d_C(\bm{x}_i, \bm{x}_j) = 1-|\text{corr}(\bm{x}_i, \bm{x}_j)|. \]

This step partitions the original $D$ features into $d \le D$ initial
groups $g_1, g_2, \ldots, g_d$.

\subsection{Screening of initial groups}
High-dimensional data are likely to contain noise features which
contribute little or even negatively to the prediction task. In such cases,
initial groups need to be screened so that noisy initial groups do not participate in the following steps. We first introduce some notation and then we present two tests for the screening of the initial groups.

\subsubsection{Notation}
We first define some notations to be used in the screening procedure. Denote by $c$ the assessment criterion of a given regression task. Typically $c$ is defined as the mean squared error (MSE) of prediction
\begin{equation} \label{eq:ac}
c = \frac{1}{N}\sum_{i=1}^N (y_i - \hat{y}_i)^2,
\end{equation}
where $\bm{y} = (y_1, \ldots, y_N)^T$ and  $\hat{\bm{y}} = (\hat{y}_1, \ldots, \hat{y}_N)^T$ are the observed values and their predictions, respectively, of the response at $N$ test points.  The data available for the application in Section~\ref{sec:CLM} allow separate test data, but usually the test points will be generated by cross validation or related methods using the training data.

To assess accuracy based on training data only, different strategies are used for the two candidate regression methods, Lasso and RF.

\begin{itemize}
\item Lasso. 

In Lasso, the predictions are produced by $K$-fold Cross-Validation (we choose $K=5$ through out the paper). More specifically, the data set $\bm{X}$ and the corresponding $\bm{y}$ are randomly grouped into $K$ folds $(\bm{X}^{(1)}, \bm{y}^{(1)})$, $\ldots$, $(\bm{X}^{(K)}, \bm{y}^{(K)})$ with $n_{(1)}, \ldots, n_{(K)}$ observations respectively ($\sum_{i=1}^K n_{(i)} = n$). Then the predictions for $\bm{y^{(i)}}$, namely $\hat{\bm{y}}^{(i)} = (\hat{y}^{(i)}_{1}, \ldots, \hat{y}^{(i)}_{n_{(i)}})^T$ is obtained by 
\[
\hat{y}^{(i)}_{j} = \hat{\bm{f}}^{(-i)}(\bm{x}^{(i)}_j),~~~~ j = 1, \ldots, n_{(i)}
\]
where $\hat{\bm{f}}^{(-i)}(\cdot)$ is the Lasso model fit from the $(K-1)$ folds other than the $i$-th fold, and $\bm{x}^{(i)}_j$ is the corresponding feature vector of $y^{(i)}_{j}$. We denote the assessment criterion $c$ for Lasso as the Cross-Validation MSE (CV-MSE).
\item Random Forests (RF).

In RF, $\hat{\bm{y}}$ can be obtained from the out-of-bag (OOB) predictions. 
The prediction $\hat{y}^{OOB}_i$ is obtained from only the trained trees that do not have $y_i$ in their bootstrapped sample, hence, the prediction is called the out-of-bag prediction. We denote the assessment criterion $c$ for RF as the out-of-bag MSE (OOB-MSE).
\end{itemize}

We further denote $\hat{\bm{y}}(g_i)$ as the vector of predictions from the base regression model (e.g.\ Lasso or RF) using only the variables in $g_i$. Let $c_i = c(\hat{\bm{y}}(g_i))$ denote the assessment measure. Denote by $\hat{\bm{y}}(g_i \cup
g_j)$ the predictions when the model is fit using the variables in $g_i$ and $g_j$ $(i \neq j)$, and denote 
\begin{equation} \label{eq:combine}
c_{ij} = c(\hat{\bm{y}}(g_i \cup g_j))
\end{equation}
as the resulting performance. Similarly, let
\begin{equation} \label{eq:ensemble}
\bar{c}_{ij} = c((\hat{\bm{y}}(g_i) + \hat{\bm{y}}(g_j)) / 2)
\end{equation}
be the performance of the ensemble of the predictions from $g_i$ and
$g_j$. 

\subsubsection{Tests for screening initial groups}
A group survives the screening if it passes the two tests described as follows. A group $g_i$ passes the first test if its own performance is ``strong'', i.e.\ high prediction accuracy. A group survives the second test if it produces ``significant combining improvement'', i.e., after combining with another group $g_j$, the model trained on the combined variables (from $g_i$ and $g_j$) produces significantly better accuracy comparing to that from $g_j$.

We use a null permutation distribution to establish the baseline for
strong individual performance and significant combining
improvement. Denote $\tilde{\bm{y}}$ as the vector of permuted response values in which
the original vector of response variable values $\bm{y}$ is randomly permuted relative
to the features. Then the counterparts of $c_i$ and $c_{ij}$ are
calculated with $\tilde{\bm{y}}$ as the response and denoted as $\tilde{c}_i$ and
$\tilde{c}_{ij}$ respectively. Denote the $\alpha$ quantile of
$\tilde{c}_i ~(i=1,\ldots,d)$ as $\tilde{p}_{\alpha}$ and the
$1-\alpha / (d-1)$ quantile of $\tilde{c}_i - \tilde{c}_{ij}~(i=1,\ldots,d;
j=1,\ldots,d)$ by $\tilde{q}_{1-\alpha / (d-1)}$. Then a group $g_i$
survives the screening if it passes both the following two tests:
\begin{enumerate}
\item $g_i$ is strong itself:
\begin{equation} \label{test1}
c_i \le \tilde{p}_{\alpha}.
\end{equation}
The rationale is that the strength of an individual initial group
should be competitive with the $\alpha$ quantile of the strengths of
initial groups with a randomly permuted response.
\item $g_i$ improves the strength of any other group $g_j$ when $g_i$ and $g_j$ are combined to build a regressor:
\begin{equation} \label{test2}
\tilde{q}_{1-\alpha / (d-1)} \le c_j - c_{ij} 
\end{equation}
The rationale is that the improvement from combining $g_i$ and $g_j$
should be competitive with the $1-\alpha / (d-1)$ quantile of combining
improvements of initial groups with a randomly permuted response. The
quantile is $1-\alpha / (d-1)$ to adjust for the $(d-1)$ tests for each initial group.  
\end{enumerate}

After the screening of initial groups, a list of surviving groups is relabeled as $\{G_1, G_2, \ldots, G_s\}$ for the next step.

\subsection{Hierarchical formation of candidate phalanxes}
We use simple greedy hierarchical clustering techniques to merge Groups $\{G_1, G_2$, $\ldots$, $G_s\}$ into phalanxes, which proves to be effective in all of our applications. Each iteration merges the pair of groups $G_i$ and $G_j$ that minimizes
\begin{equation} \label{merge}
m_{ij} = c_{ij} / \bar{c}_{ij}.
\end{equation}
Here $m_{ij} < 1$ indicates that combining $G_i$ and $G_j$ to build a single model provides more strength than ensembling two models built separately on $G_i$ and $G_j$. The number of groups, $s$, decreases by 1 after each merge. The merging process stops when $m_{ij} \ge 1$ for all $i, j$, indicating that further merging damages the performance and the resulting groups, i.e.\ $e$ candidate phalanxes $PX_1, PX_2, \ldots, PX_e$, should be now considered for ensembling.

\subsection{Screening of candidate phalanxes}
Searching for the best subset of candidate phalanxes for further ensembling is a combinatorial problem. In order to reduce the computational cost, we establish the search path $P$ in a forward selection fashion. We initialize $P$ with the candidate phalanx with the smallest value of the criterion $c$. At each stage all the remaining candidate phalanxes will be considered for ensembling with phalanxes in $P$ one at a time, and the one with the best ensembling performance with $P$ will be added to $P$. The ensembling
performance $\bar{c}_{(1,2, \ldots, h)}$ is calculated as follows. The predictions of $y_i$ from $PX_1, \ldots, PX_h$ are denoted by
$\hat{y}_{i;(PX_1)}, \ldots, \hat{y}_{i;(PX_h)}$, and the ensembled
prediction for $y_i$ is calculated as 
\begin{equation} \label{ensembleOOB}
\hat{y}_{i;(1, \ldots, h)} = \sum_{j=1}^{h} \hat{y}_{i;(PX_j)} / h
\end{equation}
The ensemble's performance is then calculated from the ensembled predictions.

For ease of description, we assume the order of entry to be  $PX_1, PX_2, \ldots, PX_k$, and the corresponding ensembling performance as each candidate is added to be $c_{PX_{\left\{1\right\}}}, c_{PX_{\left\{1, 2\right\}}}, \ldots,$ $c_{PX_{\left\{1,\ldots,k\right\}}}$ respectively. Then the set of candidate phalanxes corresponding to the best ensembling performance, say $c_{PX_{1, \ldots, h}}$, will be selected, and the remaining candidates are screened out.


After the screening of weak phalanxes, the surviving $h$ phalanxes are the
final phalanxes, and they will be ensembled in the last step.

\subsection{Ensembling of Regression Phalanxes}
We fit a model for each of the $h$ phalanxes of variables,
and obtain predictions from them. (For RF, we can increase the number of trees and get better OOB predictions as final predictions.) For a test point, the $h$ predictions from the ensemble of regression phalanxes (ERPX) are
averaged to give the final prediction.
 
\section{A simple illustrative example: Octane}
The octane data set \citep{esbensen1996multivariate} consists of NIR absorbance spectra over 226 wavelengths ranging from 1102 nm to 1552 nm with measurements every two nanometers. For each of the 39 production gasoline samples the octane number was measured. 

It is known that the octane data set contains six outliers (cases 25, 26, 36--39) to which alcohol was added. We omit those outliers to obtain 33 clean samples. 

We apply Lasso and RF separately on the data set, then we choose Lasso as the base regression model since it produces better results: the MSE of Lasso is about 0.085 compared with about 0.27 for RF.

The R package ``glmnet'' \citep{friedman2010regularization} is used for fitting Lasso models and the penalty parameter $\lambda$ is chosen by using ``cv.glmnet'' with method ``lambda.1se''. Since cross validation is used for choosing the tuning parameter $\lambda$ for Lasso, the Phalanx formation procedure has inherited randomness. Therefore, we apply both ERPX and the original Lasso to the Octane data set three times each. For ERPX, we skip the initial grouping step since the number of features is relatively small and the features are all continuous. Different numbers of surviving groups and final phalanxes are obtained. The accuracies of both methods are assessed by CV-MSE. For each run, results for mean CV-MSE over 20 CV repetitions are presented in Table~\ref{tab:octane}.

\begin{table}[h]
\centering
\small
\begin{tabular}{ccccccccccc}
  \hline
& \multicolumn{6}{c}{Number of} \\
  \cline{2-7}
&  & \multicolumn{2}{c}{Groups} & & \multicolumn{2}{c}{Phalanxes} & & \multicolumn{2}{c}{CV-MSE}\\
\cline{3-4} \cline{6-7} \cline{9-10}
 Run & Variables & Initial & Screened & & Candidate & Screened & & ERPX & Lasso \\ 
\hline
 1 & 226 & 226 & 190 & & 7 & 2 & & 0.051  & 0.084  \\ 
 2 & 226 & 226 & 195 & & 8 & 5 & & 0.049 & 0.086 \\
 3 & 226 & 226 & 192 & & 9 & 2 & & 0.044 & 0.083  \\
\hline
\end{tabular}
\caption{Number of variables, initial groups, screened groups, candidate phalanxes, screened
phalanxes and prediction accuracies for the octane data set. }
\label{tab:octane}
\end{table}

We can see that the CV-MSE values from ERPX are much smaller than those obtained from the original Lasso models, which confirms that ERPX can boost the performance of the base regression model.

In the following section, we generate synthetic data sets based on the octane data. We simulate data favoring Lasso and RF respectively as the base regression model and show that we are able to improve the performance over the base regression model using the proposed ERPX.
%

\section{Simulation studies}
In this section, we present several numerical experiments to demonstrate the performance of the proposed ERPX. We simulate data by first emulating the feature structure of the octane data. Then we generate response data as a function of the features in two different ways, to represent linear and nonlinear relationships with the features, respectively.

The data sets $\bm{X}_{(n\times p)}$ are generated based on the octane data as follows. 
\begin{itemize}
\item $\bm{X} = (\bm{x}_1, \ldots, \bm{x}_p)$, $\bm{x}_j = (x_{1j}, x_{2j}, \ldots, x_{nj})^T (j=1, \ldots, p)$, where $n=33, p=226$ as in the octane data set.
\item $\bm{X}$ is sampled from multivariate normal distribution $N(\bm{\mu}, \bm{\Sigma})$, where $\bm{\mu}$ consists of $p$ column means of the octane data. 
\item The covariance matrix $\bm{\Sigma}$ is equal to the sample covariance matrix as observed or from a perturbation of the data. Specifically, we add different levels of noise into the octane data, and for each noise level we take the resulting sample covariance matrix for $\bm{\Sigma}$. We consider three noise levels: 
\begin{itemize}
\item No noise: $\bm{\Sigma}$ is the sample covariance matrix of the octane data. 
\item Medium noise: Random samples from $N(0, 3 \sigma)$ are added to each element of the octane data, where $\sigma$ is the minimum feature standard deviation among all the 226 features of the octane data. $\bm{\Sigma}$ is then the sample covariance matrix of the modified octane data. 
\item High noise: Random samples from $N(0, 5 \sigma)$ are added to each element of the octane data. $\bm{\Sigma}$ is then the sample covariance matrix of the modified octane data. 
\end{itemize}
\end{itemize}
For each noise level we simulate the feature set $\bm{X}$ 100 times. The strategy for generating $\bm{y}$ depends on the choice of the base regression model. 
 
\subsection{Lasso as base regression model}
To favor Lasso, for each of the 100 simulated $\bm{X}$, we generate the response variable $\bm{y}$ in a linear pattern as follows.
\begin{itemize}
\item Randomly select 10 features $\bm{x}_{k_1}, \ldots, \bm{x}_{k_{10}}$ from the columns of $\bm{X}$.
\item For each feature $\bm{x}_{k_j}$ $(j=1,\ldots, 10)$, generate the corresponding coefficient $\beta_j$ from a Uniform distribution $U(0, 1)$.
\item Generate $\bm{y}_{init}$ as $\bm{y}_{init} = \sum_{j=1}^{10} \beta_j\bm{x}_{k_j}$.
\item Generate $\bm{y}$ as $\bm{y} = \min(\bm{y}_{oct}) + a\cdot \bm{y}_{init} + \bm{\epsilon}$, where $a$ is a scale constant $(\max(\bm{y}_{oct}) - \min(\bm{y}_{oct})) / (\max(\bm{y}_{init}) - \min(\bm{y}_{init}))$, $\bm{y}_{oct}$ is the original response variable from the octane data set, and $\bm{\epsilon}$ contains noises generated from $N(0, 1)$.
\end{itemize}
For each set of simulated data $(\bm{X}, \bm{y})$, we apply ERPX with base regression models Lasso and RF respectively. Since the response variable is generated as a linear combination of the predictors, Lasso is expected to perform at least as good as RF. The performance is measured by 5-fold CV-MSE and OOB-MSE for Lasso and RF respectively.

\begin{table}[h]
\centering
\small
\begin{tabular}{cccccccccc}
  \hline
& & \multicolumn{4}{c}{Average number of} \\
  \cline{3-6}
& & & \multicolumn{2}{c}{Groups} & & & \multicolumn{2}{c}{MSE}\\
\cline{4-5}  \cline{8-9}
Base & Noise & Variables & Initial & Screened & Phalanxes & & ERPX & Base \\ 
\hline
\multirow{3}{*}{Lasso} & No  & 226 & 226 & 172.32  & 6.01 & & 0.96 & 1.43  \\ 
& Medium  & 226 & 226  & 138.32 & 3.56 &  & 0.95 & 1.58  \\
& High  & 226 & 226  & 88.75 & 2.78 &  & 1.07 &  1.71    \\
\hline
\multirow{3}{*}{RF} & No  & 226 & 226 & 99.58  & 2.59 & & 0.98 & 1.32  \\ 
& Medium  & 226 & 226  & 62.93 & 2.67 &  & 0.96 & 1.40  \\
& High  & 226 & 226  & 33.35 &  2.65 &  & 1.02 & 1.56    \\
\hline
\end{tabular}
\caption{Number of variables, initial groups, screened groups, candidate phalanxes, screened
phalanxes and prediction accuracies of base regression models and ERPX for different noise levels when calculating sample covariance matrix $\bm{\Sigma}$.}
\label{tab:lassosim}
\end{table}

We can see from Table~\ref{tab:lassosim}, for all the simulation settings, regardless of the choice of the base regression model, ERPX produces more accurate predictions than the corresponding base regression model with large relative margins. Somewhat surprisingly, RF slightly outperforms Lasso in around 70\% of the 300 simulated data sets (and also on average). This could be caused by the use of different metrics, OOB-MSE versus CV-MSE, when the sample size is small.

\subsection{Random Forests as base regression model}
To generate the response variable $\bm{y}$ from highly nonlinear relationships favoring RF, we deploy the following strategy.
\begin{itemize}
\item Randomly select 10 features $\bm{x}_{k_1}, \ldots, \bm{x}_{k_{10}}$ from the columns of $\bm{X}$.
\item Generate two sets of coefficients $\bm{\beta}_1 = (\beta_{11}, \ldots, \beta_{1,10})^T$ and $\bm{\beta}_2 = (\beta_{21}, \ldots, \beta_{2,10})^T$ from a Uniform distribution $U(0, 1)$.
\item Generate  \[
	\bm{y}^{init}_{i} = \begin{cases}
			\sum_{j=1}^{10} \beta_{1j}\bm{x}_{k_j} & x_{ik_1} < \text{median}(\bm{x}_{k_1})\\
			\sum_{j=1}^{10} \beta_{2j}\bm{x}_{k_j} & x_{ik_1} \ge \text{median}(\bm{x}_{k_1})
			\end{cases}
\]
Since the initial values in $\bm{y}^{init}$ are generated from a mixture of two linear patterns, they cannot be easily modelled by linear methods such as Lasso.
\item  Generate $\bm{y} = \min(\bm{y}^{oct}) + a\cdot \bm{y}^{init} + \bm{\epsilon}$, where $a=(\max(\bm{y}^{oct}) - \min(\bm{y}^{oct})) / (\max(\bm{y}^{init}) - \min(\bm{y}^{init}))$, a scale constant, and $\bm{\epsilon}$ contains noises generated from $N(0, 1)$.
\end{itemize}
For each set of simulated data $(\bm{X}, \bm{y})$, we again apply ERPX with either Lasso or RF as the base regression model. In this case, we expect RF to outperform Lasso.
We can see from Table~\ref{tab:lassosim}, for all the simulation settings, RF outperforms Lasso as anticipated given the simulation set up, and ERPX improves upon both base regression models by large relative margins.

\begin{table}[h]
\centering
\small
\begin{tabular}{cccccccccc}
  \hline
& & \multicolumn{4}{c}{Average number of} \\
  \cline{3-6}
& & & \multicolumn{2}{c}{Groups} & & & \multicolumn{2}{c}{MSE}\\
\cline{4-5}  \cline{8-9}
Base & Noise & Variables & Initial & Screened & Phalanxes & & ERPX & Base \\ 
\hline
\multirow{3}{*}{Lasso} & No  & 226 & 226 & 150.10  & 5.29 & & 2.75 & 3.83  \\ 
& Medium  & 226 & 226  & 92.79 & 2.65 &  & 2.67 & 3.80  \\
& High  & 226 & 226  & 48.83 & 2.48 &  & 2.52 & 3.71    \\
\hline
\multirow{3}{*}{RF} & No  & 226 & 226 & 72.83  & 1.90 & & 1.59 & 2.15  \\ 
& Medium  & 226 & 226  & 37.52 & 2.04 &  & 1.66 & 2.49  \\
& High  & 226 & 226  &  16.43 &  1.86 &  & 1.57 &  2.51    \\
\hline
\end{tabular}
\caption{Number of variables, initial groups, screened groups, candidate phalanxes, screened
phalanxes and prediction accuracies of base regression models and ERPX for different noise levels when calculating sample covariance matrix $\bm{\Sigma}$.}
\label{tab:RFsim}
\end{table}

\section{Additional real data examples}
The prediction performance of ERPX is illustrated on the following data
sets.
\subsection{AID364 data set}
 Assay AID364 is a cytotoxicity assay conducted by the Scripps
Research Institute Molecular Screening Center. There are 3,286 compounds used in our study, with their inhibition percentages recorded. Visit \url{http://pubchem.ncbi.nlm.nih.gov/assay/assay.cgi?aid=364} for details. Because toxic reactions can occur in many different ways, this assay is expected to present modelling challenges. We consider five sets of descriptors for the assay, to make 5 data sets. The descriptor sets are the following: atom pairs (AP), with 380 variables; Burden numbers \citep[BN,][]{burden1989molecular}, with 24 variables; Carhart atom pairs \citep[CAP,][]{carhart1985atom}, with 1585 variables; fragment pairs (FP), with 580 variables; and pharmacophores fingerprints (PH), with 120 variables. The Burden numbers are continuous descriptors, and the other four are bit strings where each bit is set to ``1'' when a certain feature is present and ``0'' when it is not. See \citet{liu2005powermv} and \citet{hughes2010chemmodlab} for further explanation of the molecular properties captured by the descriptor sets.

The initial groups for the descriptor sets are determined by
their features names. For example, related features for FP present
similar names such as $\text{AR\_01\_AR}$, $\text{AR\_02\_AR}$, $\ldots$, $\text{AR\_07\_AR}$, and such
features will form the initial groups. We perform our proposed
ERPX on each of the five descriptor
sets. The base regression model is chosen to be RF due to its superior performance over Lasso in this case (see Table~\ref{tab:AIDS-compare} in Appendix). (RE: WILL)

ERPX and the original RF are run on each of the five descriptor sets as well as the five descriptor sets combined as a whole set, three times each. The results are presented in Table~\ref{tab:AIDS}. As we can see, ERPX provides superior prediction accuracy over the original RF, and the margin gets bigger with all five descriptor sets merged together. This is because ERPX can exploit the ``richness'' of features to improve prediction accuracy.
\begin{table}[h]
\centering
\small
\begin{tabular}{ccccccccccc}
  \hline
\multirow{3}{*}{Set} & & \multicolumn{6}{c}{Number of} \\
  \cline{3-7}
 & &  & \multicolumn{2}{c}{Groups} & & \multicolumn{2}{c}{Phalanxes} & & \multicolumn{2}{c}{OOB MSE}\\
\cline{4-5} \cline{7-8} \cline{10-11}
 & Run & Variables & Initial & Screened & & Candidate & Screened & & ERPX & RF \\ 
\hline
\multirow{3}{*}{BN} & 1 & 24 & 24 & 15 & & 4 & 4 & & 122.41 & 126.70 \\ 
& 2 & 24 & 24 & 14 & & 5 & 5 & & 123.81 & 126.97 \\
& 3 & 24 & 24 & 15 & & 5 & 5 & & 121.64 & 127.33  \\
\hline
\multirow{3}{*}{PH} & 1 & 120 & 21 & 15 & & 3 & 2 & & 131.59 & 135.70 \\ 
& 2 & 120 & 21 & 16  & & 2 & 2 & & 131.50 & 134.62 \\
& 3 & 120 & 21 & 17 & & 2 & 2 & &  131.67& 135.57 \\
\hline
\multirow{3}{*}{FP} & 1 & 580 & 105 & 64 & & 9 & 2 & & 120.03 & 125.53 \\ 
& 2 & 580 & 105 &52  & & 9 & 3 & & 120.26 & 126.15 \\
& 3 & 580 & 105 & 44 & & 6 & 2 & & 121.80 & 127.09 \\
\hline
\multirow{3}{*}{AP} & 1 & 380 & 78 & 35 & & 4 & 3 & & 124.40 & 132.07 \\ 
& 2 & 380 & 78  & 33 &  & 4 & 2 & & 124.31 & 131.73 \\
& 3 & 380 & 78 & 31& & 5 & 3 & & 124.92 & 131.69 \\
\hline
\multirow{3}{*}{CAP} & 1 & 1585 & 666 & 93 & & 11 & 2 & & 116.04 & 131.12 \\ 
& 2 &  1585 & 666  & 73 & & 9 & 2 & & 118.86 & 130.74 \\
& 3 & 1585 & 666  & 95 & & 10 & 3 & & 116.40 & 131.25 \\
\hline
\multirow{3}{*}{ALL5} & 1 & 2689 & 895 & 159 & & 21 & 6 & & 112.51 & 125.87 \\ 
& 2 & 2689 & 895 & 208 & & 20 & 5 & & 113.69 & 124.78 \\
& 3 & 2689 & 895 & 204 & & 18 & 4 & & 112.80 & 125.13 \\
\hline
\end{tabular}
\caption{Number of variables, initial groups, screened groups, candidate phalanxes, screened
phalanxes and prediction accuracies for the AID 364 assay and five descriptor sets. Three runs of ERPX are presented.}
\label{tab:AIDS}
\end{table}

Due to the naming schema of the features, the descriptor sets lend themselves well to obtaining initial groups. However, we show that ERPX still produces comparable prediction accuracies with initial groups obtained from the hierarchical clustering approach described in Section~2.1. We choose the same number of initial groups as shown in Table~\ref{tab:AIDS} to facilitate comparison. Table~\ref{tab:AIDS2} presents the new results for descriptor sets other than BN (since no initial grouping is needed). As we can see, the prediction accuracies are comparable to those in Table~\ref{tab:AIDS} where the initial grouping is done according to feature names.
\begin{table}[h]
\centering
\small
\begin{tabular}{ccccccccccc}
  \hline
\multirow{3}{*}{Set} & & \multicolumn{6}{c}{Number of} \\
  \cline{3-7}
 & &  & \multicolumn{2}{c}{Groups} & & \multicolumn{2}{c}{Phalanxes} & & \multicolumn{2}{c}{OOB MSE}\\
\cline{4-5} \cline{7-8} \cline{10-11}
 & Run & Variables & Initial & Screened & & Candidate & Screened & & ERPX & RF \\ 
\hline
\multirow{3}{*}{PH} & 1 & 120 & 21 & 14 & &  3& 2 &  & 132.06 & 135.01  \\ 
& 2 & 120 & 21 & 15  & & 3 & 2 & & 130.68 & 134.21 \\
& 3 & 120 & 21 & 15 & &3 & 2 & &  131.47& 134.78 \\
\hline
\multirow{3}{*}{FP} & 1 & 580 & 105 & 47 & & 6 & 3 & & 120.85 & 125.73 \\ 
& 2 & 580 & 105 & 44  & & 9 & 2 & & 122.85 & 125.53 \\
& 3 & 580 & 105 & 44 & & 10 & 3 & & 123.87 & 126.48 \\
\hline
\multirow{3}{*}{AP} & 1 & 380 & 78 & 20 & & 4 & 2 & & 125.23 & 132.35 \\ 
& 2 & 380 & 78  & 27 &  & 5 & 2 & & 124.02 & 131.27 \\
& 3 & 380 & 78 & 20& & 5 & 2 & & 123.95 & 132.77 \\
\hline
\multirow{3}{*}{CAP} & 1 & 1585 & 666 & 57 & & 7 & 3 & & 120.76 & 129.82 \\ 
& 2 &  1585 & 666  & 74 & & 10 & 3 & & 118.26 & 131.84 \\
& 3 & 1585 & 666  & 58 & & 7 & 5 & & 121.35 & 131.16 \\
\hline
\multirow{3}{*}{ALL5} & 1 & 2689 & 895 & 143 & & 12 & 4& & 114.15 & 125.00 \\ 
& 2 & 2689 & 895 & 89 & & 11 & 6 & & 115.46 & 123.67 \\ 
& 3 & 2689 & 895 & 74 & &  12 & 5 & & 116.44 & 125.35 \\ 
\hline
\end{tabular}
\caption{Number of variables, initial groups, screened groups, candidate phalanxes, screened
phalanxes and prediction accuracies for the AID 364 assay and four descriptor sets, with initial groups obtained from the hierarchical clustering. Three runs of ERPX are presented.}
\label{tab:AIDS2}
\end{table}

\subsection{CLM data set} \label{sec:CLM}
The community land model with carbon/nitrogen biogeochemistry (CLM-CN) is a state-of-the-art land surface model to make future climate projections. \citet{sargsyan2014dimensionality} performed simulations using CLM-CN for a single plant functional type: temperate evergreen needleleaf forest. The outputs were 100-year-later projected values of several quantities of interest (QoIs). The simulator was run 10,000 times for different settings of 79 input parameters, out of which 9983 runs succeeded.

We make predictions for two QoIs, leaf area index (LAI) and total vegetation carbon (TOTVEGC), from the 79 input variables. The two QoIs are log scaled since their sample distributions are highly right-skewed. The 9983 runs contain 38\% and $0.07$\% of zeros, respectively. Therefore, we add constants $10^{-10}$ and $10^{-13}$ to the two QoIs respectively before we apply the log scaling (these values are roughly equal to the minima of the respective non-zero values). Since the climate projections are affected by a number of uncertainties in CLM-CN, predicting the LAI and TOTVEGC is a challenging task. We choose RF as the base regression model due to its superior performance versus Lasso in this example (see Table~\ref{tab:CLM-compare} in Appendix) (RE: WILL). We apply both ERPX and the original RF to demonstrate their performance. For ERPX, we skip the initial grouping step since there are only 79 variables.

\begin{figure}[h]
\begin{center}
  \includegraphics[scale = 0.6]{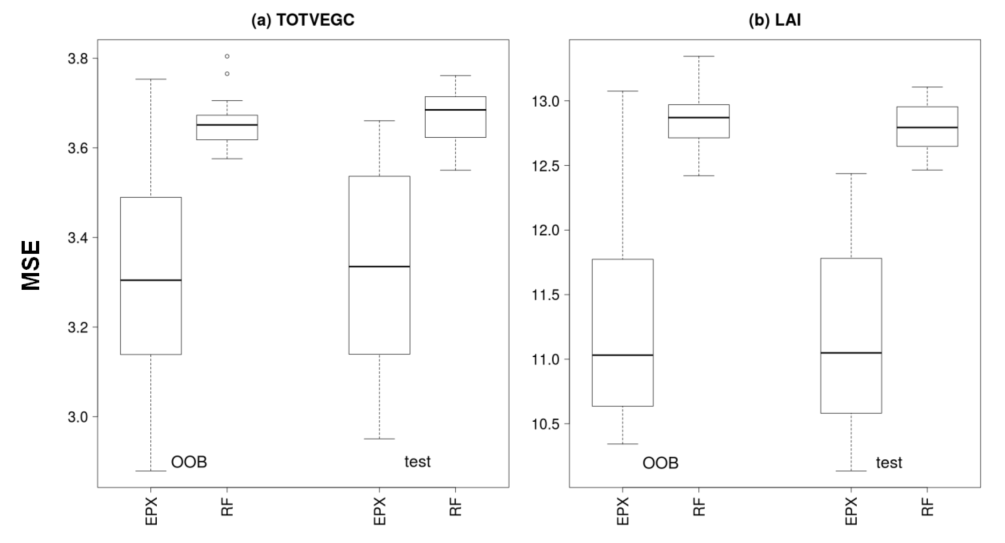}
  \end{center}
\caption{Boxplots of OOB MSE (from 5000 randomly selected training samples) and test errors (from 4983 corresponding testing samples) obtained from 20 random splits of the data into training and testing sets}
\label{fig:boxplot}
\end{figure}

The 9983 observations are randomly split into training and testing sets with 5000 and 4983 observations respectively. We repeat the random splitting 20 times to obtain 20 different pairs of training and testing sets. For each random split, both ERPX and RF are trained on the randomly sampled training set and applied to the corresponding test set. Therefore, we can generate boxplots of both the 20 OOB-MSEs from the training sets and the 20 test MSEs from the testing sets. The boxplots are shown in Figure~\ref{fig:boxplot}. It is clear that ERPX provides more accurate predictions than RF, for both TOTVECG and LAI as response variables.

We also present the detailed results from the first three splits in Table~\ref{tab:clm}. Since the number of phalanxes is always one in all the runs, the mainly difference between ERPX and RF is the screening of initial groups. By screening out most of the less important initial groups, ERPX is able to produce more accurate prediction than RF with only a small number of ``strong'' initial groups. 

\begin{table}[h]
\centering
\small
\begin{tabular}{ccccccccccc}
  \hline
\multirow{3}{*}{QoI} & & \multicolumn{6}{c}{Number of} \\
  \cline{3-7}
 & &  & \multicolumn{2}{c}{Groups} & & \multicolumn{2}{c}{Phalanxes} & & \multicolumn{2}{c}{OOB MSE}\\
\cline{4-5} \cline{7-8} \cline{10-11}
 & Run & Variables & Initial & Screened & & Candidate & Screened & & ERPX & RF \\ 
\hline
\multirow{3}{*}{TOTVEGC} & 1 & 79 & 79 & 16 & & 1 & 1 & & 3.47  & 3.66 \\ 
& 2 & 79 & 79 & 14 & & 1 & 1 & & 3.57 & 3.67 \\
& 3 & 79 & 79 & 14 & & 1 & 1 & & 3.10 & 3.62  \\
\hline
\multirow{3}{*}{LAI} & 1 & 79 & 79 & 18 & & 1 & 1 & & 10.34  & 12.63   \\ 
& 2 & 79 & 79 & 20  & & 1 & 1 & & 11.02 & 12.94 \\
& 3 & 79 & 79 & 14 & & 1 & 1 & &  11.40 & 12.89 \\
\hline
\end{tabular}
\caption{Number of variables, initial groups, screened groups, candidate phalanxes, screened
phalanxes and prediction accuracies for TOTVEGC and LAI from CLM-CN simulations. Three experiments based on random splitting of training and testing sets are presents.}\label{tab:clm}
\end{table}

\subsection{Gene expression data set}
\citet{scheetz2006regulation} conducted a study of mammalian eye diseases where the gene expressions of the eye tissues from 120 twelve-week-old male F2 rats were recorded. A gene coded as TRIM32 is of particular interest here since it is responsible for causing Bardet-Biedl syndrome.

According to \citet{scheetz2006regulation}, only 18976 probe sets exhibited sufficient signal for reliable analysis and at least 2-fold variation in expressions. The intensity values of these genes are evaluated on the logarithm scale and normalized using the method in \citet{bolstad2003comparison}. It is believed from previous studies that TRIM32 is only linked to a small number of genes, so following \citet{scheetz2006regulation} we concentrate mainly on the top 5000 genes with the highest marginal sample variance.

Again, we choose RF (MSE around 0.0128) as the base regression model over lasso (MSE around 0.0131). We apply ERPX and the original RF to this data set three times each. The results are presented in Table~\ref{tab:gene}. For ERPX, we skip the initial grouping step. 

\begin{table}[h]
\centering
\small
\begin{tabular}{ccccccccccc}
  \hline
& \multicolumn{6}{c}{Number of} \\
  \cline{2-7}
&  & \multicolumn{2}{c}{Groups} & & \multicolumn{2}{c}{Phalanxes} & & \multicolumn{2}{c}{OOB MSE}\\
\cline{3-4} \cline{6-7} \cline{9-10}
 Run & Variables & Initial & Screened & & Candidate & Screened & & ERPX & RF \\ 
\hline
 1 & 5000 & 5000 & 659 & & 4 & 2 & & 0.0112   & 0.0130   \\ 
 2 & 5000 & 5000 & 578 & & 4 & 3 & & 0.0114 & 0.0125 \\
 3 & 5000 & 5000 & 513 & & 5 & 4 & & 0.0110 & 0.0128  \\
\hline
\end{tabular}
\caption{Number of variables, initial groups, screened groups, candidate phalanxes, screened
phalanxes and prediction accuracies for the gene expression data set. }
\label{tab:gene}
\end{table}
It is clear that ERPX is providing more accurate predictions than RF for this data set.

\subsection{Glass data set}
The glass data set \citep{lemberge2000quantitative} was obtained from an electron probe X-ray
microanalysis (EPXMA) of archaeological glass samples. A total of 180 glass samples were analyzed and each glass sample has a spectrum with 640 wavelengths. The goal is to predict the concentrations of several major constituents of glass, namely, Na2O, SiO2, K2O and CaO, from the spectrum. For different responses, the choice of the base regression model varies, since neither RF nor Lasso performs uniformly better accross the response variables. We apply ERPX and the corresponding base regression model to each of the six responses three times each. The results are presented in Table~\ref{tab:glass}. As we can see, ERPX improves the prediction accuracy of the corresponding base regression method.
\begin{table}[h]
\centering
\footnotesize
\begin{tabular}{cccccccccccc}
  \hline
\multirow{3}{*}{Response} & & & \multicolumn{6}{c}{Number of} \\
  \cline{4-8}
 & & &  & \multicolumn{2}{c}{Groups} & & \multicolumn{2}{c}{Phalanxes} & & \multicolumn{2}{c}{MSE}\\
\cline{5-6} \cline{8-9} \cline{11-12}
 & Base & Run & Variables & Initial & Screened & & Candidate & Screened & & ERPX & Base \\ 
\hline
\multirow{3}{*}{Na2O} & \multirow{3}{*}{RF} & 1 & 640 & 640 & 224 & & 3 & 3 & & 0.78  & 0.94 \\ 
& & 2 & 640 & 640 & 247 & & 2 & 2 & & 0.86 & 0.92 \\
& & 3 & 640 & 640 & 245 & & 2 & 2 & & 0.85 & 0.96  \\
\hline
\multirow{3}{*}{SiO2} & \multirow{3}{*}{RF} & 1 & 640 & 640 & 86 & & 1 & 1 & & 0.98  & 1.23   \\ 
& & 2 & 640 & 640 & 86  & & 2 & 2 & & 0.98 & 1.26 \\
& & 3 & 640 & 640 & 89 & & 1 & 1 & &  0.97 & 1.22 \\
\hline
\multirow{3}{*}{K2O} & \multirow{3}{*}{Lasso} & 1 & 640 & 640 & 309 & & 1 & 1 & & 0.094 & 0.100   \\ 
& & 2 & 640 & 640 & 425  & & 1 & 1 & & 0.093 & 0.093 \\
& & 3 & 640 & 640 & 257 & & 1 & 1 & &  0.096 & 0.096 \\
\hline
\multirow{3}{*}{CaO} & \multirow{3}{*}{Lasso} & 1 & 640 & 640 & 302 & & 1 & 1 & & 0.109  & 0.111  \\
& & 2 & 640 & 640 & 201 & & 1& 1 &  & 0.109  & 0.111 \\
& & 3 & 640 & 640 & 265 & & 1 & 1 & & 0.111  & 0.112 \\
\hline
\end{tabular}
\caption{Number of variables, base regression model, initial groups, screened groups, candidate phalanxes, screened phalanxes and prediction accuracies (OOB-MSE for RF as base regression model, 5-fold CV-MSE for Lasso as base regression model) for four responses of the glass data set.}\label{tab:glass}
\end{table}

\section{Conclusion}
In this paper, we propose a novel framework called ensemble of Regression Phalanxes (ERPX). We propose to divide a often large number of features into subsets called Regression Phalanxes. Separate predictive models are built using features in each Regression Phalanx and they are further ensembled. 

The proposed approach is widely applicable. We have demonstrated it on a variety of applications spanning chemistry, drug discovery, climate-change ecology, and gene expression. The simulated examples and real applications demonstrate that ERPX can take advantage of the richness of the features and produce gains in prediction accuracy over effective base regression model such as Lasso and RF. 

\section{Appendix}
\begin{table}[h]
\centering
\small
\begin{tabular}{ccc}
  \hline
 \multirow{2}{*}{Set} &  \multicolumn{2}{c}{MSE}\\
 \cline{2-3}
  & Lasso & RF\\
  \hline
  BN & 152.53 & 127.39 \\
  PH & 145.29 & 135.01 \\
  FP & 143.56 &  125.98 \\
  AP & 148.45 & 131.76\\
  CAP & 146.57 & 131.67 \\
  \hline
\end{tabular}
\caption{MSEs of Lasso (5-fold CV-MSEs) and RF (OOB-MSEs) for five descriptor sets of the AID 364 assay}
\label{tab:AIDS-compare}
\end{table}

\begin{table}[h]
\centering
\small
\begin{tabular}{ccc}
  \hline
 \multirow{2}{*}{Response} &  \multicolumn{2}{c}{MSE}\\
 \cline{2-3}
  & Lasso & RF\\
  \hline
LAI  & 12.37 & 11.58 \\
TOTVEGC & 3.47 & 3.29 \\
  \hline
\end{tabular}
\caption{MSEs of Lasso (CV-MSEs) and RF (OOB-MSEs) for two responses of the CLM data}
\label{tab:CLM-compare}
\end{table}

\newpage

\bibliographystyle{apalike}
\bibliography{rshc}

\end{document}